\documentclass{InUSTrans}

\usepackage[dvipsnames]{xcolor}
\usepackage{times}
\usepackage{soul}
\usepackage{url}
\usepackage{amsmath}
\usepackage{amsthm}
\usepackage{booktabs}
\usepackage{algorithm}
\usepackage{algpseudocode}
\usepackage[switch]{lineno}
\usepackage{amssymb}
\usepackage{longtable}
\usepackage{booktabs} 
\usepackage{multirow}
\usepackage{subcaption}  
\usepackage{tabularx}
\journalvolumn{xx}
\journalnumber{x}
\journalyear{xxxx}
\setarticlestartpagenumber{1}
\journaltype{}

\allowdisplaybreaks

\begin{document}
\title{Halo Separation-guided Underwater Multi-scale Image Restoration}

\author{
Jiaxin Yang$^{1*}$, 
Honglin Liu$^{2}$,
Yongli Wang$^{1}$, 
Shuyi Cao$^{1}$,
Chengcheng Jiang$^{1}$,
Jiale Wang$^{1}$
}

\address{
$^{1}$ College of Information Science and Technology, Dalian Maritime University, Dalian, 116000, China 
\\
$^{2}$ College of Marine Electrical Engineering, Dalian Maritime University, Dalian, 116000, China \\
$^{*}$ Corresponding author.
}

\begin{abstract}
Underwater images captured by Autonomous Underwater Vehicles (AUVs) are inevitably affected by artificial light sources, which often produce halos in the foreground of the camera and seriously interfere with the quality of the image. The existing underwater image enhancement methods fail to fully consider this key problem, and the robustness of processing images under artificial light scenes is poor. In practical applications, since underwater image enhancement itself is a very challenging task, the influence of artificial light sources will lead to serious degradation of image performance and affect subsequent vision tasks. In order to effectively deal with this problem, this paper designs a single halo image correction method based on an iterative structure. The network is mainly divided into two sub-networks, one is the halo layer separation sub-network which aims to separate the halo by gradient minimization, and the other is the multi-scale recovery sub-network which aims to recover the image information masked by halo. The UIEB and EUVP synthetic datasets are used for training to ensure that the network can fully learn the characteristics and laws of underwater halo images. Then a large number of halo images taken in an underwater environment with real artificial light are collected for testing. In addition, the brightness distribution characteristics of underwater halo images are analyzed and the radial gradient is introduced to constraint eliminate halo to improve the effect of underwater image restoration. 
\end{abstract}

\begin{keywords}
Underwater halo image; radial gradient; residual network; artificial light sources
\end{keywords}

\maketitle


\section{INTRODUCTION}
In recent years, propelled by the decreasing marginal utility of global terrestrial resources and the escalating requirements for deep-sea strategic resource exploitation, the demand for Autonomous Underwater Vehicles (AUVs) in marine scientific exploration, inspection, and salvage domains has witnessed a consistent upward trend. Nevertheless, owing to the distinctive optical attenuation properties of aqueous media, the performance of the vision systems integrated with AUVs is substantially constrained, leading to the prevalent issue of poor visibility during underwater camera imaging. To enable clear perception of the surrounding underwater environment, the deployment of artificial lighting devices for ambient light compensation has become a common practice. While this approach effectively enhances the luminance of the target area, the combined influence of light sources and suspended particulates often induces the formation of halos in the foreground of captured images, as illustrated in Fig. 1. Such visual artifacts pose significant challenges to various computer vision applications, including image stitching, object recognition, and depth estimation, all of which rely on precise and consistent scene intensity measurements for accurate operation \cite{y2}. 
\begin{figure}[h]
    \centering
    \includegraphics[width=\columnwidth,height=3cm]{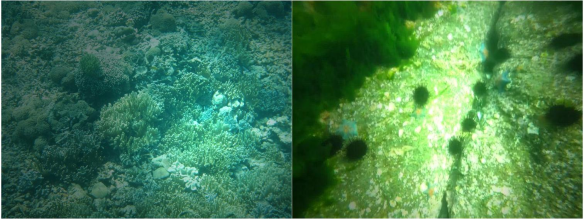}
    \caption{Underwater vignetting image.}
    \label{fig:1}
    \vskip -1.5pc
\end{figure} 

The current mainstream underwater image restoration algorithms mostly focus on visibility restoration under uniform illumination conditions \cite{18,25,26}. In practice, when dealing with local high intensity artificial light, the enhanced image can lead to severe performance degradation. Therefore, although how to effectively address the halo phenomenon caused by artificial light sources mounted on AUVs has rarely been considered, its significance is significant. Typically, underwater environments are often characterized by low-illumination conditions because light propagating in water is subject to wavelength and range-dependent absorption and scattering. To address such optical attenuation effects and improve environmental visibility, searchlights are widely installed above or on both sides of the AUVs to compensate for illumination. Unfortunately, when the artificial light sources form a high-intensity spot in the near-field target area, it will form a unique halo effect in the camera foreground. Specifically, the illumination intensity distribution of the underwater image has a significant radial attenuation from the center of the light sources. 

This luminance inhomogeneity makes it difficult to analyze the characteristics of image edge regions or dark corner regions, because the effective visual information is limited. To deal with the challenge of halo effect caused by artificial light sources, this paper proposes an underwater image restoration method based on iterative structure of single image halo correction. It is considered that the distribution characteristics of artificial light sources in underwater images have similar radial attenuation characteristics to the vignetting of lenses and CCD components \cite{y1}. Based on this, an improved underwater imaging model combining the characteristics of artificial light sources is proposed to promote the generalization ability of underwater artificial light scenes. In order to effectively avoid the problem of over-enhancement and over-exposure caused by the existing underwater image enhancement (UIE) methods when dealing with data with suboptimal illumination conditions, the radial gradient constraint is used to separate the halo layer from the non-uniformly illuminated underwater halo image. The main contributions of this paper are summarized as follows:

\begin{itemize}
\item To address the detrimental impact of halos induced by the deployment of artificial light sources, this paper presents an iterative-based single-image halo correction method, specifically designed for the restoration of underwater halo-affected images. The proposed method comprises two key components: a halo layer separation sub-network and a multi-scale restoration sub-network.

\item In the halo layer separation network, motivated by the resemblance between the brightness distribution of the dark corner in the foreground of terrestrial camera images and that of underwater artificial light sources, the radial gradient theory is incorporated. This incorporation enables the network to prioritize the separation of the halo layer.

\item Following the separation of the halo layer from the underwater halo-affected image, although the illumination intensity distribution becomes uniform, color fidelity and fine-scale details are still compromised. To address this issue, a multi-scale restoration network is devised to recover the image's color characteristics and detailed information.

\end{itemize}

\section{RELATED WORK}
Underwater image enhancement methodologies \cite{daiti5,daiti13} can be systematically categorized into three distinct groups: physical model-based approaches, non-physical model-based techniques, and deep learning-driven methods. These enhancement strategies \cite{daiti14} serve as crucial technical underpinnings for subsequent tasks, including underwater target detection and robotic grasping operations.

\subsection{Physical Models-based}
Model-based underwater image enhancement methods primarily rely on Jaffe's underwater imaging model, which involves estimating the transmission map and background light. Most of these methods are based on Dark Channel Prior (DCP) \cite{y3}. To adapt DCP to underwater scenarios, various optimized variants have been proposed. For instance, Yuan et al. \cite{y4} developed a multi-scale fusion enhancement method to restore image texture details. Some solutions focusing on the depth of underwater scenes were also introduced, and Fu et al. \cite{y5} used the homology between the original and degraded underwater images to achieve underwater image enhancement. Imposing homology constraints between the underwater original image and the redescended image is equivalent to minimizing the recovery error, which can be used for image restoration. Due to the lack of light in the water, the quality of underwater images is seriously degraded. Although artificial lighting facilitates underwater imaging, it often induces non-uniform illumination. To address this, Hou et al. \cite{y6} proposed an Illumination Channel Sparsity Prior (ICSP)-guided variational framework for restoring underwater images with uneven lighting. They also devised a fast numerical algorithm based on the alternating direction method of multipliers \cite{1,3,4} to expedite the optimization process \cite{10,12}. While these model-based approaches demonstrate notable performance in specific scenarios, their reliance on prior knowledge undermines their robustness in real-world underwater image enhancement.

\subsection{Non-physical Models-based}
The aim of the non-physical model approaches is to directly adjust the image pixel values in order to improve the visual quality of underwater images. For example, based on Retinex decomposition \cite{y7}, automatic white balance \cite{y8}, etc. Zhuang et al. \cite{y9} introduced a Retinex-inspired model leveraging super-Laplacian reflection prior for underwater image enhancement. The approach employs an alternating minimization algorithm \cite{6,7,8,9,10} to restore true color naturalness by imposing norm penalties on both the first and second-order gradients of the reflectance, thereby establishing a super-Laplacian reflectance prior. Mi et al. \cite{y10} proposed a new multi-purpose oriented underwater image enhancement method. Firstly, the input image is decomposed into illumination layer and reflection layer. Then, the brightness of the illumination layer is compensated, and the color correction and contrast enhancement of the reflection layer are performed by multi-scale processing strategy. To address underwater image quality degradation, Zhang et al. \cite{y11} proposed an attenuation map-guided color correction strategy to rectify color distortion. The method applied a global contrast strategy via maximum information entropy optimization for color-corrected images and introduced a weighted wavelet visual perception fusion strategy to integrate high/low-frequency components across scales, yielding high-quality underwater images. Although the non-physical model method improves the visual quality of underwater images, it does not consider the optical characteristics of underwater imaging, which will make the color of the image lose the sense of reality, and the effectiveness of processing underwater halo images is reduced.

\subsection{Deep Learning-based}
In recent years, deep learning methods \cite{y12, 11, 16} have made breakthroughs in computer vision. However, limited by the difficulty of obtaining real images under different water quality conditions, many existing methods mainly rely on the learning of the mapping relationship between synthetic underwater images and corresponding clear images. To simulate underwater degraded images that are closer to the real scene, researchers have proposed generative models such as FUnIE-GAN \cite{y13} and UWCNN \cite{y14}. To enhance the generalization capability of data-driven methods in real-world scenarios, Li et al. \cite{y15} constructed the first real paired underwater image dataset for deep network training and designed Water-Net, an underwater enhancement network. Experimental results validate the dataset's effectiveness in improving model generalization performance. Peng et al. \cite{y16} were the first to introduce the Transformer model into the underwater image enhancement task, proposing a network model referred to as U-shape. This network incorporated a channel-based multi-scale feature fusion module and a spatial global feature modeling module, both specifically designed to address the challenges of the enhancement task. Since few people consider the impact of the illumination distribution of the original image on the enhancement results, Li et al. \cite{y17} designed an underwater image enhancement network guided by a brightness mask with multi-attention embedding to enhance the dark regions while restraining the excessive few algorithms suitable for underwater robots using artificial light sources, which will cause the result of overexposure of halo regions when processing underwater images carrying halo.

\subsection{Discussion}
Most existing methods\cite{jia2,jia4} focus on general underwater image enhancement\cite{jia3} or non-uniform illumination correction, while the halo phenomenon caused by artificial light sources in underwater environments remains insufficiently explored. 
Moreover, existing vignetting\cite{jia1} or illumination correction methods typically rely on polynomial fitting, physical illumination assumptions or global illumination adjustment strategies, which are often inadequate for modeling the complex, spatially varying halo artifacts induced by artificial light sources.

In contrast, our method targets the more complex halo phenomenon caused by artificial light sources in underwater environments, which exhibits strong spatial variability and cannot be well modeled by conventional vignetting assumptions. 
To address this, we introduce a radial-gradient-constrained iterative learning mechanism to explicitly separate the halo layer, enabling structured decomposition rather than implicit correction. Combined with a multi-scale recovery network, the proposed method effectively restores both illumination uniformity and fine details.

\section{METHOD}
In existing research, land-based halo image correction methods are widely applied in related fields. However, significant differences between underwater and terrestrial environments render terrestrial halo correction methods ineffective for underwater halo images, resulting in suboptimal correction outcomes. An ideal underwater halo image correction method should effectively eliminate halo-induced degradation, adaptively adjust illumination in dark regions, correct color casts, and enhance visibility. In this paper, an underwater halo image correction driven network is constructed to solve the problem. Firstly, the implementation details and evaluation of the proposed framework are presented. In the following, we detail the key components of our approach. A single halo image correction method based on iterative structure proposed in this paper includes two modules. The first module is a halo layer separation sub-network, and the second module is a multi-scale recovery sub-network.

\subsection{Overall Network Structure}
Inspired by the terrestrial halo image imaging model, an underwater halo image imaging model is constructed. It is assumed that the underwater halo image can be decomposed into two components: a low-quality underwater image with uniform illumination and a halo layer as shown in the Eq \eqref{fomula1}.
\begin{equation}\begin{aligned}
\label{fomula1}
{Z_{low}} = {I_{low}} \bullet {v_{low}}
\end{aligned}\end{equation}
where $Z_{low}$ represents the underwater halo image and $I_{low}$ represents the low-quality underwater image with uniform illumination, enhancement of the bright regions. $v_{low}$ is the halo layer, which is affected by the introduction of artificial lighting, and $\bullet$ is the multiplication of elements one by one. Despite the remarkable progress of underwater enhancement methods, there are still problems such as color and detail loss.

\begin{figure*}[t]
    \centering
    \includegraphics[width=15cm,height=7cm]{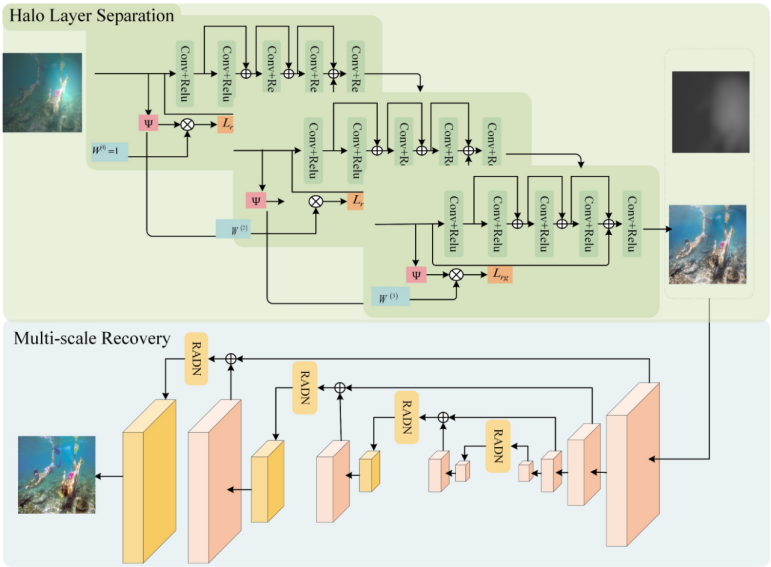}
    \caption{Network structure diagram.}
    \label{fig:2}
    \vskip -1.5pc
\end{figure*} 

Therefore, in this paper we propose a novel method, mainly to eliminate the halo and enhance the color and details of the underwater halo image. Inspired by the above model, a single underwater halo image correction network based on iterative structure is designed to realize the recovery of underwater halo images, and the overall network structure is shown in Fig. \ref{fig:2}.

\subsection{Halo Layer Separation Network}
The halo layer separation network is used to separate the halo layer from underwater halo images, thereby obtaining low-quality underwater images without halos. Inspired by iterative minimization\cite{17,19} of sparse gradients and radial gradients, we construct an iterative network architecture to extract the halo layer from halo images and design a novel objective function to constrain the network's training process. The objective function is based on the analysis of the radial gradient characteristics of underwater halo images. By constructing the model expression reasonably and introducing it into the iterative network for training, the halo layer in the halo image can be effectively separated. The objective function of the design is shown as Eq. \eqref{fomula2}:
\begin{equation}\begin{aligned}
\label{fomula2}
v = \arg \min ||{v_{gt}} - {v_{low}}|{|_2} + \lambda ||(|\Psi ({v_{gt}}) - \Psi (v_{low}^k)|)|{|_1}
\end{aligned}\end{equation}
where $\Psi$ represents the radial gradient, i.e. the radial gradient of the pixels in the image relative to the center of the image light source. Its expression is shown in Eq. \eqref{fomula3}:
\begin{equation}\begin{aligned}
\label{fomula3}
\Psi (x,y) = \frac{{\nabla v(x,y) \bullet \mathop r\limits^ \to  (x,y)}}{{|\mathop r\limits^ \to  (x,y)|}}
\end{aligned}\end{equation}
where $\mathop r\limits^ \to  (x,y) = {[x - {x_0},y - {y_0}]^T}$ represents the vector along the radial direction formed by the pixel ${(x,y)}$ and the center of the light source $({x_0},{y_0})$. ${\nabla v(x,y)}$ represents the gradient of the halo layer $v$ and $ \bullet $ represents the inner product \cite{2,6,9}.

It is worth mentioning that in practice, underwater imaging systems may involve multiple artificial light sources, especially in AUV scenarios with lights mounted at different positions. This can lead to more complex illumination patterns than a single radial distribution. 
To address this, we approximate the halo structure using a dominant illumination center, which can be estimated from the brightest region or the intensity centroid of the image. This approximation is based on the observation that, even under multi-light configurations, the illumination field is often dominated by a primary region or can be represented by an effective unified center.
Furthermore, it should be noted that the proposed radial gradient constraint serves as a structural prior rather than a strict physical assumption. Therefore, the model does not require a perfect single-center condition and remains robust to moderately complex or multi-source illumination patterns.

Inspired by the iterative reweighted least squares method to solve the non-convex optimization subproblem \cite{7,8}, radial gradient constrained iterative weight regularization is introduced to improve the gradient sparsity of the CNN output image. Eq. \eqref{fomula2} is transformed into the following iterative optimization process:
\begin{equation}\begin{aligned}
\label{fomula4}
  &{v^k} = \arg \min ||{v_{gt}} - {v_{low}}|{|_2}  \\
  &\hspace{4mm}+\lambda ||{w^k}(|\Psi ({v_{gt}}) - \Psi (v_{low}^k)|)|{|_1} \hfill \\
  &{w^k} = \frac{1}{{\Psi ({v_{gt}}) - \Psi (v_{low}^{k - 1}) + \varepsilon }} \hfill \\
  &{w^1} = 1 \hfill \\ 
\end{aligned}\end{equation}
The above formulation employs the iterative reweighted least squares (IRLS) technique, which views the optimization process as a series of standard least squares problems. In each of these problems, a weighting factor is utilized, which is determined according to the solution of the prior iteration.
$w \odot {v_{gt}} $ denotes the real halo layer generated by matlab and $v_{low}$ represents the halo layer separated from the network. $\Psi ({v_{gt}})$ and $\Psi ({v_{low}})$ represent the radial gradients of the true halo layer and the separated halo layer, respectively. $k$ indicates the $k$-th iteration. In this paper, an iterative network is built to solve Eq. \eqref{fomula4} step by step. And further slight changes are made by setting progressively increasing to promote image sparsity.

According to Eq. \eqref{fomula4}, in this paper, a fully convolutional network is constructed to decompose the halo layer and low-quality underwater images. The halo layer is separated from the input underwater halo image through a progressive residual network, and the network is trained under the guidance of Eq. \eqref{fomula4}. The halo layer separated from the network should be smooth and not contain too many details. Therefore, we introduce the smoothing loss:
\begin{equation}\begin{aligned}
\label{fomula5}
  {L_{smooth}} = {\lambda _1}||\nabla {v_{gt}} - \nabla {v_{low}}|{|_2} 
\end{aligned}\end{equation}
where $\lambda _1$ denotes the weight coefficient and ${\nabla v}$ represents the gradient of the halo layer. The smoothing loss is introduced as an auxiliary regularization term and is jointly optimized with the objective in Eq. \eqref{fomula4} during network training. After obtaining the halo layer ${v_{low}}$, a low-quality underwater image ${I_{low}}$ with uniform illumination can be obtained by element-wise division, as shown in Eq. \eqref{fomula6}.
\begin{equation}\begin{aligned}
\label{fomula6}
  {I_{low}} = \frac{{{Z_{low}}}}{{{v_{low}}}}
\end{aligned}\end{equation}

\subsection{Multi-scale Recovery Network}

\begin{figure}[t]
    \centering
    \includegraphics[width=\columnwidth]{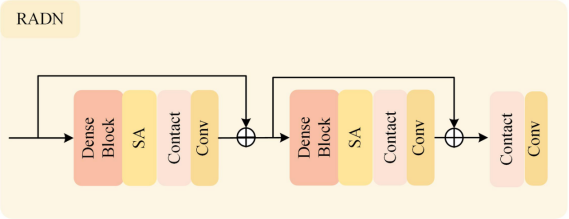}
    \caption{RADN module}.
    \label{fig:3}
    \vskip -1.5pc
\end{figure}

The low-quality underwater images without halos obtained after the halo layer separation sub-network still have problems, such as color and detail loss. Therefore, the focus in this section is to improve the quality of underwater images with uniform illumination without halos. Specifically, a multi-scale recovery sub-network is proposed. By using a multi-scale residual dense network as the backbone network, multi-scale information can be aggregated to obtain a wider range of information, thereby better restoring the color and details of low-quality underwater images without halos. The used Residual Attention Dense Network (RADN) is shown in Fig. \ref{fig:3}, which is mainly composed of Residual Dense Blocks (RDBS) stacked. In addition, the channel attention mechanism is introduced to better recover the color and details of the image. Global residual and global feature fusion are added to the network back-end, and the size of the remaining convolution layer is set to 3×3 except that the size of the convolution layer for local and global feature fusion is 1×1. In the multi-scale recovery sub-network, the image features are downsampled to different scales (1/4, 1/2 and 1), and then upsampled gradually from small scale to large scale, and fused before processing.

The process of eliminating the halo may damage the color and details of the image. Therefore, a recovery sub-network is employed to enhance the details and color fidelity of the halo-free underwater images obtained from the halo layer separation network. In this recovery sub-network, this paper proposes the reconstruction loss $L_{re}$ to constrain and enhance the global feature differences between underwater images and real images, and to guide the direction of gradient descent in network training.
\begin{equation}\begin{aligned}
\label{fomula7}
  {L_{re}} = {\mu _1}||{I_{pre}} - {I_{normal}}|{|_2}
\end{aligned}\end{equation}
where $I_{pre}$ is the underwater image recovered by the multi-scale restoration sub-network, and $I_{normal}$ is the underwater reference image.

This paper further designs a joint loss to constrain the detail and color enhancement of the image. To be more specific, to minimize the structural discrepancies between the predicted underwater image and the ground-truth underwater image, the SSIM loss is introduced, ensuring that the resulting underwater image aligns better with human visual perception. Considering that the noise and detail features are mainly reflected in the gradient domain, the gradient loss was also designed to ensure the consistency between the restored underwater image and the reference image. The specific formula is as follows:
\begin{equation}\begin{aligned}
\label{fomula8}
& {L_{pre}} =  - {\mu _2}{\text{SSIM(}}{I_{pre}},{I_{normal}}{\text{)}} \hfill \\
&{L_{rg}} = {\mu _3}||\Psi ({I_{pre}}) - \Psi ({I_{normal}})|{|_2} \hfill \\ 
\end{aligned}\end{equation}

Therefore, the overall loss function is formulated as a combination of the above three components as Eq. \eqref{fomula9}:
\begin{equation}\begin{aligned}
\label{fomula9}
L_{\text{recovery}} = L_{re} + L_{pre} + L_{rg}
\end{aligned}\end{equation}
where $\mu_1$, $\mu_2$, and $\mu_3$ are trade-off parameters that balance the contributions of reconstruction fidelity, structural similarity, and gradient consistency, respectively.

\section{EXPERIMENTS}
To verify the effectiveness of the network, we conduct a series of experiments. In this section, we first discuss the experimental setup of our study, followed by the experimental results analysis including comparison experiments, evaluation experiments and ablation experiments.

\begin{figure}[t]
    \centering
    \includegraphics[width=\columnwidth,height=4cm]{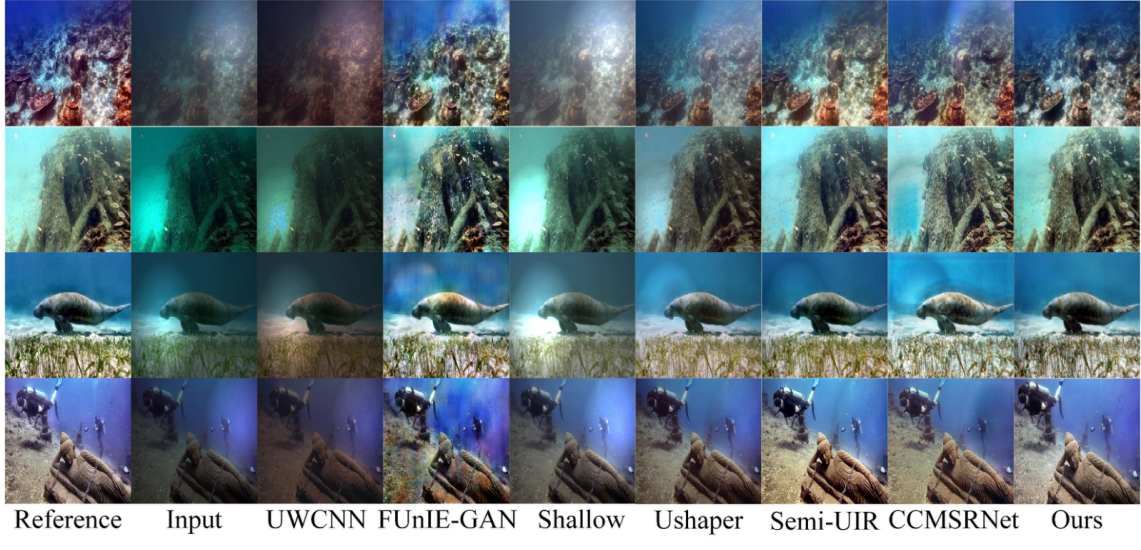}
    \vskip -0.5pc
    \caption{Subjective visual comparison on the UIEB. }
    \label{fig:4}
    \vskip -1pc
\end{figure} 
\begin{figure}[t]
    \centering
    \includegraphics[width=\columnwidth,height=3.5cm]{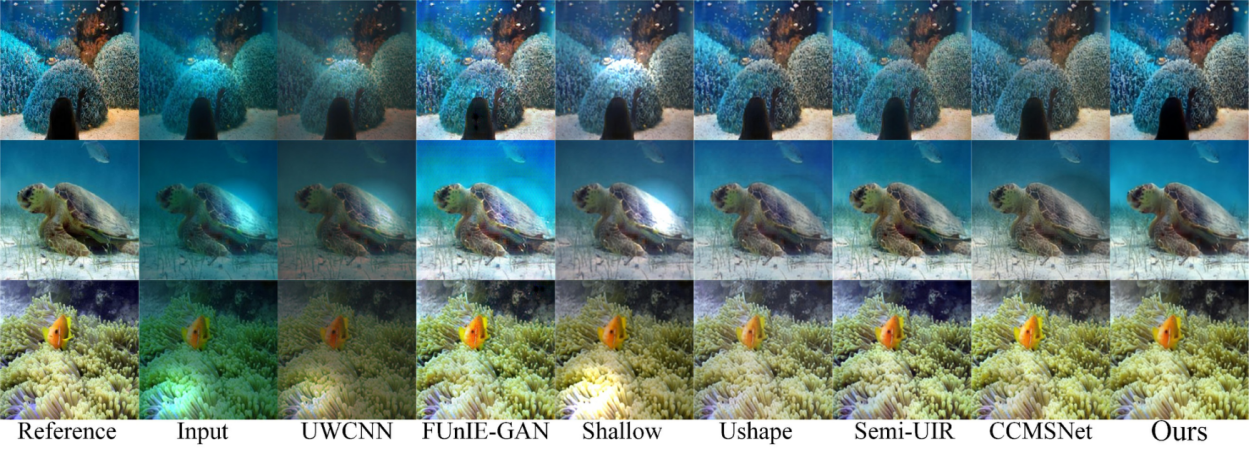}
    \vskip -0.5pc
    \caption{Subjective visual comparison on the EUVP.}
    \label{fig:5}
    \vskip -1.5pc
\end{figure} 
\begin{figure}[t]
    \centering
    \includegraphics[width=\columnwidth,height=3.5cm]{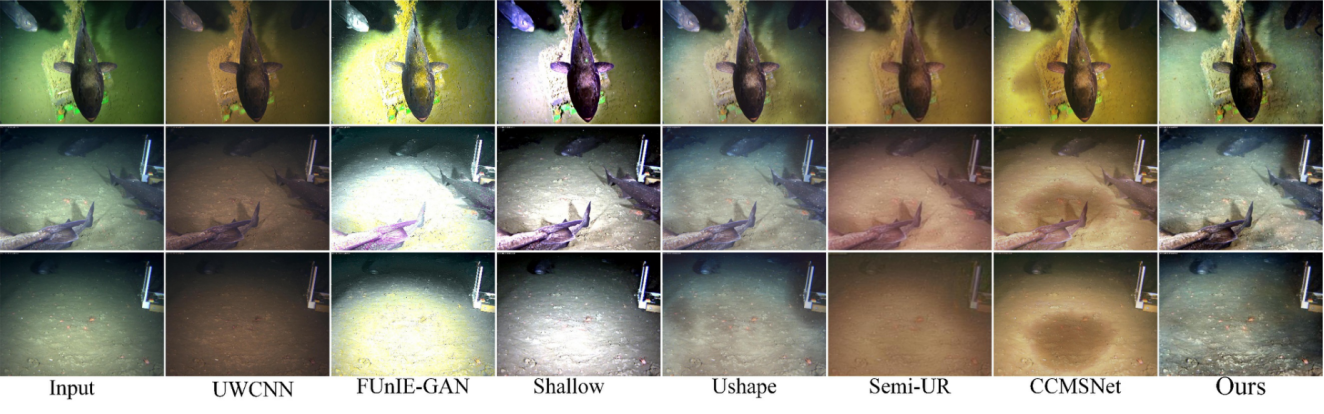}
    \vskip -0.5pc
    \caption{Subjective visual comparison on the OceanDark.}
    \label{fig:6}
    \vskip -1.5pc
\end{figure}

\begin{table}[t]
\centering
\caption{Reference Evaluations on UIEB Dataset}
\label{tab1}
\footnotesize
\renewcommand{\arraystretch}{1.2}
\begin{tabular}{c|c|c|c|c}
\hline
\textbf{Method} & \textbf{SSIM$\uparrow$} & \textbf{MSE$\downarrow$} & \textbf{PSNR$\uparrow$} & \textbf{PCQI$\uparrow$} \\ 
\hline
UWCNN     & 0.588 & 5.583 & 12.938 & 0.351 \\
FUnIE     & 0.728 & 1.251 & 20.196 & 0.717 \\
Shallow   & 0.632 & 2.974 & 15.726 & 0.384 \\
U-shape    & 0.767 & 1.276 & 19.947 & 0.603 \\
Semi-UIR  &\underline{0.807} &0.694 & 21.076 & 0.707 \\
CCMSNet   & 0.801 & \underline{0.687} &\underline{21.463} &\underline{0.761} \\
\hline
\textbf{Ours} & \textbf{0.814} & \textbf{0.662} & \textbf{22.929} & \textbf{0.806} \\
\hline
\end{tabular}
\end{table}

\subsection{Dataset Introduction}
In order to fully verify the fairness of the experiment and the generalization ability of the model, this paper uses a batch of underwater image datasets with halo effect constructed based on two public real underwater scene datasets of UIEB and EUVP for training and testing. 

Since real underwater images with paired halo-free and halo-degraded versions are difficult to obtain, we generate synthetic halo-degraded images to form supervised training pairs. Specifically, for each reference image, a halo layer is generated to simulate the illumination effect of artificial light sources in underwater environments. The halo layer is constructed based on a radial attenuation pattern centered at a selected light-source position, where the intensity gradually decreases from the center to the periphery. The generated halo layer is then applied to the reference image via element-wise multiplication to produce the degraded input image. In this way, paired data of halo-degraded image, reference image are obtained for training.

The constructed datasets are adapted to the lighting scenes of real AUV operations. The use of these datasets is described as follows. The UIEB \cite{y15} dataset is divided into training set and test set, which contain 800 images and 90 images respectively. 3700 images from the underwater scene subset of the EUVP \cite{y14} dataset. Of these images, 3500 images are used for training and 200 for testing. In addition, OceanDark \cite{y19} dataset taken under real artificial light environment is used for testing. 
To be more specific, the model evaluated on the OceanDark dataset is trained using only the synthetic halo-degraded data constructed from the UIEB and EUVP datasets. No OceanDark images are used during training. The trained model is directly applied to the OceanDark dataset in order to evaluate the generalization ability of the proposed method under real-world underwater lighting conditions. This setting ensures a fair and unbiased evaluation of cross-dataset generalization performance.

Although synthetic halo layers are used for training, the diversity of their spatial distribution and intensity helps approximate real underwater illumination conditions. To evaluate real-world applicability, we conduct experiments on the OceanDark dataset, which contains images captured under real artificial lighting conditions with complex and irregular halo effects. More importantly, evaluation on the OceanDark dataset demonstrates that the proposed method generalizes well to real-world halo patterns that deviate from the synthetic model. This suggests that the model learns a generalized representation of halo structures rather than merely fitting the synthetic generation process.

\subsection{Experimental Environment}
The experiment is based on Pytorch framework, the network model is built in python3.8 environment, and Adam optimizer is used for training. Among them, the learning rate is 0.001, the patch size is 48, the batch size is 2, and the training is performed on the RTX2070Ti server with 12G video memory size.

During comparison experiments, we do not directly use publicly available pretrained models of other methods, since they are typically trained on different datasets and settings, which may lead to unfair comparisons under our constructed halo-degraded dataset. To ensure a fair comparison, all baseline methods are re-trained by us under the same training setup and data configuration as the proposed method. Specifically, we use the same training datasets, data splits, and preprocessing procedures for all compared methods.

\subsection{Results Analysis}

\begin{table}[t]
\centering
\caption{Reference Evaluations on EUVP Dataset}
\label{tab2}
\footnotesize
\renewcommand{\arraystretch}{1.2}
\begin{tabular}{c|c|c|c|c}
\hline
\textbf{Method} & \textbf{SSIM$\uparrow$} & \textbf{MSE$\downarrow$} & \textbf{PSNR$\uparrow$} & \textbf{PCQI$\uparrow$} \\ 
\hline
UWCNN     & 0.638 & 3.660 & 15.334 & 0.334 \\
FUnIE     & 0.758 & \underline{0.159} & \underline{23.451} & 0.634 \\
Shallow   & 0.644 & 1.394 & 18.942 & 0.432 \\
U-shape    & 0.737 & 1.058 & 22.009 & 0.575 \\
Semi-UIR  &\underline{0.816} &0.266 & 22.996 & 0.676 \\
CCMSNet   & 0.803 & 0.163 &23.350 &\underline{0.756} \\
\hline
\textbf{Ours} & \textbf{0.827} & \textbf{0.151} & \textbf{23.683} & \textbf{0.786} \\
\hline
\end{tabular}
\end{table}

\begin{table}[t]
\centering
\caption{No-reference evaluation on UIEB dataset}
\label{tab3}
\footnotesize
\renewcommand{\arraystretch}{1.2}
\begin{tabular}{c|c|c|c}
\hline
\textbf{Method}  & \textbf{Entropy$\uparrow$} & \textbf{UIQM$\uparrow$} & \textbf{UCIQE$\uparrow$} \\ 
\hline
UWCNN     & 6.5017 & 1.2651 & 0.5517  \\
FUnIE     & 7.1365 & 1.4710 & 0.5969 \\
Shallow   & 7.3426 & 1.1937 & 0.5645  \\
U-shape    & 7.3905 & 1.3621 & 0.5712  \\
Semi-UIR  &7.6065 &\underline{1.5042} & \underline{0.6133}  \\
CCMSNet   & \underline{7.6388} & 1.4612  &0.5996 \\
\hline
\textbf{Ours} & \textbf{7.7182}  & \textbf{1.5120} & \textbf{0.6693} \\
\hline
\end{tabular}
\end{table}
\begin{table}[t]
\centering
\caption{ No-reference evaluation on EUVP dataset}
\label{tab4}
\footnotesize
\renewcommand{\arraystretch}{1.2}
\begin{tabular}{c|c|c|c}
\hline
\textbf{Method}  & \textbf{Entropy$\uparrow$} & \textbf{UIQM$\uparrow$} & \textbf{UCIQE$\uparrow$} \\ 
\hline
UWCNN     & 6.7783 & 1.2225 & 0.5229  \\
FUnIE     & 7.0463 & 1.4298 & 0.5873 \\
Shallow   & 7.1349 & 1.2973 & \underline{0.5933}  \\
U-shape    & 7.3868 & 1.3072 & 0.5578  \\
Semi-UIR  &7.2839 &1.4442 & 0.5727  \\
CCMSNet   & \underline{7.6045} & \textbf{1.5212}  &0.5806 \\
\hline
Ours & \textbf{7.6997}  & \underline{1.5052} & \textbf{0.6171} \\
\hline
\end{tabular}
\end{table}
\begin{table}[t]
\centering
\caption{ No-reference evaluation on OceanDark dataset}
\label{tab5}
\footnotesize
\renewcommand{\arraystretch}{1.2}
\begin{tabular}{c|c|c|c}
\hline
\textbf{Method}  & \textbf{Entropy$\uparrow$} & \textbf{UIQM$\uparrow$} & \textbf{UCIQE$\uparrow$} \\ 
\hline
UWCNN     & 6.4489 & 0.8734 & 0.4960  \\
FUnIE     & 6.6216 & 0.9452 & 0.5434 \\
Shallow   & 6.5376 & 0.8845 & \underline{0.5855}  \\
U-shape    & 6.6783 & \underline{1.2228} & 0.5095  \\
Semi-UIR  &\underline{7.1004} &1.1514 & 0.5774  \\
CCMSNet   & 6.3273 & 1.1529  &0.5845 \\
\hline
Ours & \textbf{7.3347}  & \textbf{1.3458} & \textbf{0.6058} \\
\hline
\end{tabular}
\end{table}

\begin{table}[t]
\centering
\caption{Reference evaluation indicators for different ablation experiment results}
\label{tab6}
\footnotesize
\renewcommand{\arraystretch}{1.2}
\begin{tabular}{c|c|c|c|c}
\hline
\textbf{Method} & \textbf{SSIM$\uparrow$} & \textbf{MSE$\downarrow$} & \textbf{PSNR$\uparrow$} & \textbf{PCQI$\uparrow$} \\ 
\hline
w/o IRLS       & 0.784 & 0.976 & 20.914 & 0.781 \\
w/o smooth     & 0.828 & 0.974 &21.896 & 0.789 \\
w/o RE         & 0.819 & 1.098 & 20.626 & 0.794 \\
w/o v-net      & 0.855 & 0.894 & 21.937 & 0.809 \\
w/o R-net      &0.838  & 1.051& 20.676 & 0.797 \\
\hline
Ours & \textbf{0.874} & \textbf{0.792} & \textbf{22.929} & \textbf{0.828} \\
\hline
\end{tabular}
\end{table}

\begin{table}[t]
\centering
\caption{No-reference evaluation indicators for different ablation experiment results}
\label{tab7}
\footnotesize
\renewcommand{\arraystretch}{1.2}
\begin{tabular}{c|c|c|c}
\hline
\textbf{Method}  & \textbf{Entropy$\uparrow$} & \textbf{UIQM$\uparrow$} & \textbf{UCIQE$\uparrow$} \\ 
\hline
w/o IRLS    & 7.3241 & 1.4283 & 0.5793  \\
w/o smooth     & 7.2114 & 1.4314 & 0.6072 \\
w/o RE   & 7.6126 & 1.4186 & 0.5533  \\
w/o v-net    & 7.3856 & 1.4072 & 0.5991  \\
w/o R-net  &7.1193 &1.4442 &0.5727  \\
\hline
Ours & \textbf{7.7312 }  & \textbf{1.5922} & \textbf{0.6263} \\
\hline
\end{tabular}
\end{table}

In this paper, six underwater image enhancement methods based on deep learning are selected for comparative tests, namely FUnIE-GAN \cite{y13}, UWCNN \cite{y14}, U-shape \cite{y16}, Shallow \cite{y20}, Semi-UIR \cite{y21}, CCMSRNet \cite{y22}. Figs. \ref{fig:4}- \ref{fig:6} show the visual comparison results on UIEB, EUVP and OceanDark datasets using different methods. 
The following observations are obtained:

(1) In these figures, the input image represents the underwater halo image in the artificial light source environment. The image itself has details loss and the introduction of artificial light sources causes halo, resulting non-uniform illumination and other problems. This seriously affects the quality of underwater images. Moreover, it can be seen from the figures that the performance of most underwater image enhancement methods is limited to varying degrees when dealing with underwater images with halo taken in an artificial light source environment. 
The comparison of the result shows that the halo area is prone to over-exposure when UWCNN, FUnIE-GAN and Shallow methods process underwater halo images. UIESS method has a certain effect but there are still problems of detail loss and color deviation. The Ucolor, Semi-UIR and CCMSR-Net methods have achieved good results in processing halo images, but the existence of halo can still be seen from the resulting images. The U-shape method works well in dealing with halos, but the recovery of detail loss due to halos is not as good as the proposed method.

(2) The proposed networks are compared on test sets (UIEB, EUVP), and since these test sets contain truth-valued data, four full-reference metrics are used to measure the performance of these methods. Commonly used full reference evaluation metrics are Structural Similarity (SSIM) Peak Signal-to-Noise Ratio (PSNR), Mean Squared Error (MSE), and Patch-based Contrast Quality Index (PCQI). Tables \ref{tab1} and \ref{tab2} list the comparison of the full reference evaluation indexes of the proposed method and other deep learning methods on some randomly selected result graphs on the UIEB and EUVP dataset, respectively, where the optimal and suboptimal values are marked in bold and underlined.

In Table \ref{tab1}, the proposed method achieves the best results in the four evaluation indicators on the two dataset, which proves that the results obtained by our method are closer to the reference image.
Table \ref{tab2} shows that the MSE evaluation index of the proposed method is worse than that of the CCMSNet method, but the visual comparison results show that the results obtained by the CCMSNet method still have halo phenomenon. The optimal values of the two evaluation indexes of signal-to-noise ratio and image quality are achieved, and the results obtained by our method are more in line with human visual perception.

(3) In order to better prove the robustness of the proposed method, we also compare the no-reference evaluation indicators on three test sets (UIEB, EUVP, OceanDark). Because OceanDark does not contain real data, the performance can only be measured using no-reference metrics for comparison. Commonly used underwater no-reference Image Quality assessment indicators include Entropy, Universal Image Quality Measure (UIQM) and Unified Color Image Quality Evaluator (UCIQE).

The quantitative analysis results in Table \ref{tab3} show that the no-reference evaluation indexes of the results obtained by the proposed method on the UIEB dataset all achieve optimal values. It proves that the results obtained by the proposed method achieve good results both in detail feature preservation and color restoration.
Table \ref{tab4} shows that the proposed method also performe well on the EUVP dataset. Although the UIQM evaluation index achieves suboptimal value, the optimal result is achieved in the Entropy evaluation index, which indicates that the result graph obtained by our method retains more detailed information. At the same time, the best results are also obtained on the color evaluation index UCIQE, which proves that the results obtained by our method have achieved excellent results. Meanwhile, as shown in Table \ref{tab5}, the three indicators Entropy, UIQM, and UCIQE on OceanDark dataset achieve optimal values.

(4) It should be noted that entropy, while indicative of pixel intensity distribution, does not directly measure true information content or physical fidelity. High entropy values may also arise from increased contrast or edge sharpness.
In our method, the entropy improvement is mainly attributed to the removal of halo-induced illumination bias and the recovery of structural details, rather than artificial sharpening. The proposed radial gradient constraint and smoothing regularization ensure that the recovered details remain consistent with the underlying image structure.

Furthermore, the division operation is stabilized by the smoothness constraint on the estimated halo layer, which reduces the risk of numerical amplification. The subsequent recovery network further refines the results to avoid unrealistic artifacts.
The effectiveness of our method on real-world datasets such as OceanDark also suggests that the enhancement is not limited to synthetic effects, but generalizes to complex real-world imaging conditions.

Overall, results on the UIEB, EUVP and OceanDark datasets demonstrate that the proposed method can not only solve the halo problem in the image, but also achieve good results in the recovery of details.

\subsection{Ablation Study}
In this section, the ablation experiments\cite{15,20,21}, the contribution of the proposed objective function and core modes to the network will be analyzed. 

Different components in the underwater halo image correction network are excluded one by one to prove the overall influence of this part on the proposed model. The specific experimental content includes: verifying the effectiveness of the proposed iterative objective function, the effectiveness of smoothing loss, reconstruction loss, the effectiveness of halo layer separation sub-network, and the effectiveness of multi-scale recovery sub-network.
The quantitative evaluation results on UIEB are shown in Tables \ref{tab6} and \ref{tab7}. It can be seen that our method achieves the optimal values in each objective evaluation index. 

In summary, from the perspective of quantitative data of objective evaluation, both the proposed loss function and the network module demonstrate significant effectiveness.

\section{CONCLUSION}
Aiming at the problem that the pixel intensity in the near light area is large and the pixel intensity in the far away area is low, which is caused by artificial light sources, this paper proposes a single underwater halo image correction method based on iterative structure. The network consists of a halo layer separation sub-network and a multi-scale recovery sub-network. The former is used to extract the halo of the underwater halo image, and the latter is used to correct the color and enhance the details of the low-quality underwater image after extracting the halo. Since the network is trained in an end-to-end form based on deep learning, it is more robust than traditional methods for underwater image enhancement. This method not only effectively solves the halo problem caused by artificial light sources, but also improves the image contrast and restores the natural color of the image.

\section*{CONFLICT OF INTEREST}
The authors declare no conflicts of interest.








\begin{reference}

\bibitem{y2} J. Jaffe, “Computer modeling and the design of optimal underwater imaging systems,” \textit{IEEE Journal of Oceanic Engineering}, vol. 15, no. 2, pp. 101–111, 1990.

\bibitem{18} G. Jiang et al., “Visual in-context
learning for underwater image restoration,” \textit{IEEE Signal Processing}
Letters, vol. 33, pp. 1072–1076, 2026.

\bibitem{25} G. Ju et al., “A lightweight polarization-guided plug-in for underwater
image enhancement,” \textit{IEEE Transactions on Circuits and Systems for
Video Technology}, vol. 36, no. 3, pp. 2729–2743, 2026.

\bibitem{26} Y. Wang et al., “Multi-parameter detection based on u-shaped biomimetic optical fiber sensor,” \textit{Journal
of Lightwave Technology}, vol. 43, no. 16, pp. 7954–7963, 2025.

\bibitem{y1} B. Yu et al., “Vignetting correction using an optical model and constant chromaticity prior,” \textit{IEEE Transactions on Computational Imaging}, vol. 9, pp. 1071–1083, 2023.



\bibitem{y3} P. L. Drews et al., “Underwater depth estimation and image restoration based on single images,” \textit{IEEE Computer Graphics and Applications}, vol. 36, no. 2, pp. 24–35, 2016.

\bibitem{y4} J. Yuan et al., “Tebcf: Real-world underwater image
texture enhancement model based on blurriness and color fusion,” \textit{IEEE
Transactions on Geoscience and Remote Sensing}, vol. 60, pp. 1–15,
2022.


\bibitem{y5} Z. Fu et al., “Unsupervised underwater image restoration: From a homology perspective,”
\textit{Proceedings of the AAAI Conference on Artificial Intelligence}, vol. 36,
no. 1, pp. 643–651, Jun. 2022.



\bibitem{y6} G. Hou et al., “Non-uniform illumination underwater image restoration via illumination channel sparsity
prior,” \textit{IEEE Transactions on Circuits and Systems for Video Technology},
vol. 34, no. 2, pp. 799–814, 2024.


\bibitem{y7} Y.-H. Lin and Y.-C. Lu, “Low-light enhancement using a plug-and-play retinex model with shrinkage mapping for illumination estimation,”
\textit{IEEE Transactions on Image Processing}, vol. PP, pp. 1–1, 07 2022.



\bibitem{y8} C. O. Ancuti et al., “Color
balance and fusion for underwater image enhancement,” \textit{IEEE Transactions on Image Processing}, vol. 27, no. 1, pp. 379–393, 2018.



\bibitem{y9} P. Zhuang et al., “Underwater image enhancement
with hyper-laplacian reflectance priors,” \textit{IEEE Transactions on Image
Processing}, vol. 31, pp. 5442–5455, 2022.


\bibitem{y10} Z. Mi et al., “A
vignetting-correction-based underwater image enhancement method for
auv with artificial light,” \textit{IEEE Journal of Oceanic Engineering}, vol. 50,
no. 1, pp. 213–227, 2025.


\bibitem{y11} W. Zhang et al.,
“Underwater image enhancement via weighted wavelet visual perception fusion,” \textit{IEEE Transactions on Circuits and Systems for Video Technology}, vol. 34, no. 4, pp. 2469–2483, 2024.


\bibitem{y12} X. Liu et al., “Mda-net: A multi-distribution aware network for underwater image enhancement,” \textit{IEEE
Transactions on Geoscience and Remote Sensing}, vol. 63, pp. 1–13, 2025.


\bibitem{y13} M. J. Islam et al., “Fast underwater image enhancement
for improved visual perception,” \textit{IEEE Robotics and Automation Letters},
vol. 5, no. 2, pp. 3227–3234, 2020.


\bibitem{y14} C. Li et al., “Underwater scene prior inspired
deep underwater image and video enhancement,” \textit{Pattern Recognition},
vol. 98, p. 107038, 2020. 


\bibitem{y15} C. Li et al., “An
underwater image enhancement benchmark dataset and beyond,” \textit{IEEE
Transactions on Image Processing}, vol. 29, pp. 4376–4389, 2020.


\bibitem{y16} L. Peng et al., “U-shape transformer for underwater image enhancement,” \textit{IEEE Transactions on Image Processing}, vol. 32,
pp. 3066–3079, 2023.


\bibitem{y17} Y. Li et al., “Underwater image enhancement via brightness mask-guided multi-attention embedding,” \textit{Signal Processing}:
Image Communication, vol. 130, p. 117200, 2025.

\bibitem{y19} T. P. Marques and A. Branzan Albu, “L2uwe: A framework for the efficient enhancement of low-light underwater images using local contrast
and multi-scale fusion,” in 2020 \textit{IEEE/CVF Conference on Computer
Vision and Pattern Recognition Workshops (CVPRW)}, 2020, pp. 2286–
2295.


\bibitem{y20} A. Naik et al., “Shallow-uwnet: Compressed model for underwater image enhancement (student abstract),” \textit{Proceedings of the AAAI Conference on Artificial Intelligence}, vol. 35, no. 18, pp. 15 853–15 854, May 2021.

\bibitem{y21} S. Huang et al., “Contrastive semi-supervised learning for underwater image restoration via reliable bank,”
in \textit{2023 IEEE/CVF Conference on Computer Vision and Pattern Recog-
nition (CVPR)}, 2023, pp. 18 145–18 155.


\bibitem{y22} H. Qi et al., “Deep color-corrected multiscale retinex network for underwater image enhancement,” \textit{IEEE Transactions
on Geoscience and Remote Sensing}, vol. 62, pp. 1–13, 2024.


\bibitem{1} J. Yang et al., “Parameter-free multi-view clustering via refined tensor
learning,” \textit{Neurocomputing}, vol. 656, p. 131497, 2025.


\bibitem{2} J. Peng et al., “Bi-level inter-modality
modulation for unsupervised visible-infrared person re-identification,”
\textit{IEEE Transactions on Information Forensics and Security}, vol. PP, pp.
1–1, 01 2026.

\bibitem{3} H. Wang et al., “Tensor
completion framework by graph refinement for incomplete multi-view
clustering,” \textit{IEEE Transactions on Multimedia}, vol. 27, pp. 9385–9398,
2025.

\bibitem{4} H. Wang et al., “Graph-collaborated auto-encoder hashing for multiview binary clustering,” \textit{IEEE Transactions on Neural Networks and Learning Systems}, vol. 35, no. 7, pp. 10 121–10 133, 2024.


\bibitem{daiti5} H. Zhang et al., “Amplitude -
phase decomposition-based latent diffusion model for underwater image enhancement,” \textit{Expert Systems with Applications}, vol. 312, p. 131296, 2026.


\bibitem{6} J. Peng et al., “Adaptive memorization with group
labels for unsupervised person re-identification,” \textit{IEEE Transactions
on Circuits and Systems for Video Technology}, vol. 33, no. 10, p.
5802–5813, Oct. 2023.



\bibitem{7} M. Yao et al., “Between/within view information completing for tensorial incomplete multi-view clustering,” \textit{IEEE
Transactions on Multimedia}, vol. 27, pp. 1538–1550, 2025.

\bibitem{8} Q. Liu et al.,
“Consensus-guided incomplete multi-view clustering via cross-view affinities learning,” in \textit{Proceedings of the Thirty-Fourth International Joint Conference on Artificial Intelligence, IJCAI-25}, International Joint Conferences on Artificial Intelligence Organization, 8 2025, pp. 5761–5769, main Track.

\bibitem{9} B. Cai et al., “Focus more on what? guiding multi-task training for end-to-end person search,” \textit{IEEE Transactions on
Circuits and Systems for Video Technology}, vol. 35, no. 7, pp. 7266–
7278, 2025.



\bibitem{10} H. Wang et al.,
“Manifold-based incomplete multi-view clustering via bi-consistency guidance,” \textit{IEEE Transactions on Multimedia}, vol. 26, pp. 10 001–
10 014, 2024.


\bibitem{11} J. Peng et al. “Omni contextual aggregation
networks for high-fidelity image inpainting,” \textit{IEEE Transactions on
Circuits and Systems for Video Technology}, vol. 35, no. 6, pp. 6129–
6144, 2025.


\bibitem{12} G. Jiang et al., “Reliable feature
imputation with cross-view relation transfer for deep incomplete multi-view classification,” \textit{IEEE Transactions on Circuits and Systems for
Video Technology}, pp. 1–1, 2026.


\bibitem{daiti13} R. Zeng et al.,
“Amplitude exchanging network for unsupervised underwater image enhancement,” \textit{Pattern Recognition}, vol. 175, p. 113119, 2026.


\bibitem{daiti14} G. Fan et al.,
“Dcd-uie: Decoupled chromatic diffusion model for underwater image enhancement,” \textit{IEEE Transactions on Image Processing}, vol. 35, pp.
449–464, 2026.



\bibitem{15} J. Peng et al., “Refid: Reciprocal frequency-
aware generalizable person re-identification via decomposition and filtering,” \textit{ACM Trans. Multimedia Comput. Commun. Appl.}, vol. 20, no. 7,
Apr. 2024.

\bibitem{16} Z. Sun et al., “Hierarchical sequential context modelling for high-fidelity image inpainting,” \textit{IEEE Transactions on
Circuits and Systems for Video Technology}, pp. 1–1, 2025.


\bibitem{17} G. Jiang et al., “Hybrid anchor
graph learning and tensorized spectral embedding fusion for multi-view
clustering,” \textit{Neurocomputing}, vol. 673, p. 132768, 2026.


\bibitem{19} J. Yang et al.,
“Hierarchical structure-guided incomplete multi-view tensor clustering,”
\textit{Applied Intelligence}, vol. 56, no. 4, Feb. 2026.

\bibitem{20} H. Wang et al., “Tensorized parameter-free multi-view spectral clustering based on fair representation learning,” \textit{IEEE Transactions on Multimedia}, pp. 1–13, 2026.

\bibitem{21} J. Tan et al., “Unsupervised lifelong person re-identification via affinity harmonization,” \textit{ACM Trans.
Multimedia Comput. Commun. Appl.}, vol. 22, no. 4, Mar. 2026.

\bibitem{jia1}K. Sooknanan et al., “Improving underwater visibility using vignetting correction,” in \textit{Visual Information Processing and Communication III}, vol. 8305, International Society for Optics and Photonics. SPIE, 2012, p. 83050M.

\bibitem{jia2}Y. Li et al., “Underwater image devignetting and colour correction,” in \textit{Image and Graphics}. Cham: Springer International Publishing, 2015, pp. 510–521.

\bibitem{jia3}J. Li et al., “Watergan: Unsupervised generative network to enable real-time color correction of monocular underwater images,” \textit{IEEE Robotics and Automation Letters}, vol. PP, 02 2017.

\bibitem{jia4} Y. Wang et al., “Underwater vignetting image correction based on binary polynomial regularization and latent low-rank representation,” \textit{IEEE Transactions on Circuits and Systems for Video Technology}, vol. 35, no. 4, pp. 3410–3425, 2025.


\end{reference}
\end{document}